\makeatletter
\@ifundefined{iffull}{%
	\newif\iffull
	\fulltrue
}{}
\@ifundefined{ifshort}{%
	\newif\ifshort
	\shortfalse
}{}
\makeatother

\iffull
\documentclass{article}

\usepackage{amsfonts, amsthm, amsmath, amssymb,hyperref}
\usepackage{xcolor}

\usepackage{fullpage}

\usepackage[style=alphabetic,natbib=true,backend=bibtex,maxbibnames=20,maxcitenames=6,giveninits=true,doi=false,url=true]{biblatex}
\providecommand*{\citet}[1]{\AtNextCite{\AtEachCitekey{\defcounter{maxnames}{2}}} \textcite{#1}}

\bibliography{vf-allrefs-local,privstab}

\theoremstyle{plain}
\newtheorem{theorem}{Theorem}
\newtheorem{lemma}{Lemma}

\newtheorem{corollary}{Corollary}
\newtheorem{remark}{Remark}

\else

\documentclass[final]{colt2020} 

\usepackage{times}

\fi

\usepackage{bbm}

\newcommand{\flipk}{\mathrm{flip}(k)}

\newcommand{\sta}{\gamma}

\renewcommand{\phi}{\varphi}

\renewcommand{\epsilon}{\varepsilon}

\newcommand{\I}{\mathcal{I}}
\newcommand{\R}{\mathbb{R}}

\renewcommand{\eqref}[1]{Eq.~(\ref{#1})}

\newcommand{\A}{\mathcal{A}}
\newcommand{\VC}{\mathrm{VC}}

\renewcommand{\tilde}[1]{\widetilde{#1}}
\providecommand{\dfn}{:=}

\usepackage{vfmacros}

\newif\ifnotes
\notestrue

\ifnotes
\newcommand{\yuval}[1]{\textcolor{red}{\textbf{Yuval: {#1}}}}
\newcommand{\vitaly}[1]{\textcolor{blue}{\textbf{Vitaly: {#1}}}}
\else
\newcommand{\yuval}[1]{}
\newcommand{\vitaly}[1]{}
\fi

\iffull
\author{Yuval Dagan\thanks{Part of the work was done while the author was at Google Research.}\\ MIT
	 \and Vitaly Feldman\thanks{Now at Apple. Part of this work was done while the author was visiting the Simons Institute for the Theory of Computing.} \\ Google Research}
\else
\coltauthor{%
 \Name{Vitaly Feldman}\thanks{Now at Apple. Part of this work was done while the author was visiting the Simons Institute for the Theory of Computing.}\\
 \addr Google Research
 \AND
 \Name{Yuval Dagan}\thanks{Part of the work was done while the author was at Google Research.}\\
 \addr MIT
}
\fi

\title{PAC learning with stable and private predictions}
\begin{document}
	\maketitle

\ifshort
\begin{abstract}
	We study binary classification algorithms for which the prediction on any point is not too sensitive to individual examples in the dataset. Specifically, we consider the notions of uniform stability (Bousquet and Elisseeff, 2001) and prediction privacy (Dwork and Feldman, 2018). Previous work on these notions shows how they can be achieved in the standard PAC model via simple aggregation of models trained on disjoint subsets of data. Unfortunately, this approach leads to a significant overhead in terms of sample complexity. Here we demonstrate several general approaches to stable and private prediction that either eliminate or significantly reduce the overhead. Specifically, we demonstrate that for any class $C$ of VC dimension $d$ there exists a $\gamma$-uniformly stable algorithm for learning $C$ with excess error $\alpha$ using $\tilde O(d/(\alpha\gamma) + d/\alpha^2)$ samples. We also show that this bound is nearly tight. For $\eps$-differentially private prediction we give two new algorithms: one using $\tilde O(d/(\alpha^2\eps))$ samples and another one using $\tilde O(d^2/(\alpha\eps) + d/\alpha^2)$ samples. The best previously  known  bounds for these problems are $O(d/(\alpha^2\gamma))$ and $O(d/(\alpha^3\eps))$, respectively.
	\end{abstract}

\section{Introduction}




For a domain $X$ and $Y=\zo$, let $\A$ be a learning algorithm that given a dataset $S\in (X\times Y)^n$ and a data-point $x \in X$ outputs a randomized prediction $\A^S(x)$. (Equivalently, one can think of algorithm $\A$ as returning a randomized classifier $\A^S:X\to \zo$). We consider algorithms $\A$ that for every $x,y$, $\pr[\A^S(x) = y]$ is not sensitive to individual examples in $S$. More formally, we say that $\A$ is an $(\eps,\delta)$-differentially private prediction algorithm \citep{DworkFeldman18} if for any pair of datasets $S$ and $S'$ that differ in a single example and every $(x,y)\in X\times Y$ we have
$$\pr_{\A}[\A^S(x)=y] \leq e^\eps \cdot \pr_{\A}[\A^S(x)=y] + \delta .$$
The setting when $\eps=0$ is equivalent to the $\delta$-uniform stability\footnote{Alternatively, uniform stability can be considered for algorithms that output a model $h: X\to [0,1]$ that gives the confidence level at $x$. It is easy to see \citep{ElisseeffEP05}, that our discussion and results extend to this setting by defining $h(x) \dfn \E[\A^S(x)]$.}\citep{BousquettE02}.

Stability is a classical approach for understanding and analysis of generalization bounds \citep{RogersWagner78,DevroyeW79,BousquettE02,ShwartzSSS10,FeldmanV:19}. In practice, stability-inducing methods such as Bagging \citep{breiman1996bagging} and regularization \citep{BousquettE02,ShwartzSSS10} are used to improve accuracy and robustness to outliers.

Prediction privacy was defined by \citet{DworkFeldman18} to study privacy-preserving learning in the setting where the description of the model trained on sensitive data is not accessible to the (potentially adversarial) users. Instead, the users have access to the model through an interface (or API) which is provided by the data curator. A query to this interface is an input point  $x\in X$ and, in response, the user receives a prediction $y\in Y$ on $x$ of the model. Thus the definition of prediction privacy is exactly the differential privacy of the output of the curator on a single query measured with respect to the training data.

Many of the existing ML systems, such as query prediction in online search and credit scores, are deployed in this way.  At the same time, as demonstrated in a growing number of works, even such black-box exposure of a learned model presents significant privacy risks. For example, for models trained using typical deep learning algorithms, such access allows to infer whether a point $x$ is present in the training dataset with high accuracy \citep{ShokriSSS17,LongBG17,truex2018towards} and to complete sentences used in the training data \citep{carlini2019secret}. Thus understanding of prediction privacy is a crucial step towards ensuring that such systems preserve the privacy of the users who contributed the training data. \citet{BassilyTT:18} consider the same notion while focusing on algorithms for answering multiple queries to the curator by aggregating multiple non-private learning algorithms.

For comparison, the standard setting of privacy-preserving learning aims to ensure that the model learned from the data is produced in a differently private way. Thus this approach preserves privacy even when a potential adversary has complete access to the description of the predictive model. The downside of this strong guarantee is that for some learning problems, achieving the guarantee is known to have substantial additional costs, both in terms of sample complexity and computation. In particular, such guarantee may be an overkill in the setting where only the predictive interface needs to be exposed.

Another application of algorithms with private predictions is for labeling public unlabeled data. Training on data labeled in this way gives a differentially private learning algorithm \citep{HammCB16,PapernotAEGT17,papernot2018scalable,BassilyTT:18}. It is also easy to see that these notions of stability ensure some level of protection against targeted data poisoning attacks \citep{Biggio:2012} in which an attacker can add examples to the dataset with the goal of changing the prediction on a point they choose.

Given the significant benefits of these notions of stability and privacy it is natural to ask what is the overhead of ensuring them. We will address this question in the classical context of learning a class of Boolean functions in the (agnostic) PAC learning framework \citep{Valiant:84,Haussler:92,KearnsSS:94}. Namely, for a class $H$ of Boolean functions over $X$, we say that $\A$ is an $(\alpha,\beta)$-agnostic PAC learning algorithm for $H$ if for every distribution $D$ over $X\times Y$, given a dataset $S$ of i.i.d.~examples from $D$, $\A$ outputs a hypothesis $h$ that with probability at least $1-\beta$ over the choice of $S$ satisfies:
$$\pr_{(x,y)\sim D, \A}[\A^S(x) \neq y] \leq \inf_{f\in H} \pr_{(x,y)\sim D}[f(x) \neq y] + \alpha .$$
Namely, the {\em excess error} is at most $\alpha$. In the realizable setting it is additionally known that there exists $f\in H$ such that for all $(x,y)$ in the support of $D$, $y= f(x)$.

A simple, general, and well-known approach to improve stability is via averaging of models trained on disjoint subsets of data. Alternatively, one can pick a random subset of the dataset and run the algorithm on that subset. Clearly, this technique improves stability by a factor $k$ at the expense of using $k$ times more data. It was used in an influential work of \citet{ShwartzSSS10} to demonstrate that learnability of $H$ implies existence of a uniformly stable algorithm for learning $H$ and, to the best of our knowledge, this is the only known technique for learning an arbitrary class $H$ of finite VC dimension with uniform stability. An unfortunate disadvantage of this technique is that it substantially increases the sample complexity. In particular, it uses $O(d/(\gamma\alpha^2))$ samples to learn a class of VC dimension $d$ with uniform stability $\gamma$ and excess error $\alpha$ (the dependence on $\beta$ is logarithmic and thus we set it to a small fixed constant). By the same argument, in the realizable case $O(d/(\gamma\alpha))$ samples suffice.

A somewhat more careful aggregation and analysis are needed to achieve privacy since naive averaging does not improve the $\eps$ parameter. In the realizable case $\tilde O(d/(\alpha\eps))$ samples are known to suffice (and this is tight), but in the agnostic case the best known bound was $\tilde O(d/(\alpha^3\eps))$ \citep{DworkFeldman18,BassilyTT:18}. An additional factor of $\alpha$ results from the need to add some noise in the aggregation step while ensuring that it contributes at most $\alpha$ to the excess error.

For comparison, we note that ensuring that the entire model is produced with differential privacy cannot, in general, be done using a number of samples polynomial in $d$. For example, threshold functions on a line have VC dimension of $1$ but require an infinite number of samples to learn with privacy \citep{FeldmanXiao15,AlonLMM19}. The only general approach for PAC learning with differential privacy is the technique of \citet{KasiviswanathanLNRS11} that is based on the exponential mechanism \citep{McSherryTalwar:07}. This approach leads to sample complexity of $O(\log(|H|)/(\alpha\eps) + d/\alpha^2)$. In this bound $\log(|H|)$ can be replaced with the {\em representation dimension} \citep{BeimelNS:13} of $H$ which can sometimes be lower than $\log(|H|)$.

\subsection{Our contribution}
We describe several new algorithms for agnostic PAC learning of arbitrary VC classes that significantly reduce the overheads of achieving stability and privacy of predictions. We first show a simple and natural algorithm that is $\gamma$-uniformly stable and is a $(\alpha,\beta)$-agnostic PAC learner with the nearly optimal sample complexity of $\widetilde{O}(d/(\gamma\alpha)+d/\alpha^2)$.
	\begin{theorem} \label{thm:main-stability}
For every class $H$ of VC dimension $d$ and $\alpha,\beta,\gamma \in (0,1)$, there exists a $\gamma$-uniformly stable $(\alpha,\beta)$-agnostic PAC learning algorithm with sample complexity of
		\[
		n \le O\left( \frac{d \log(1/\alpha)}{\gamma \alpha}
		+ \frac{d + \log(1/\beta)}{\alpha^2}\right).
		\]
	\end{theorem}
This bound is tight up to poly-logarithmic factors. First, a lower bound of $n = \Omega((d+\log(1/\beta))/\alpha^2)$ holds for any $(\alpha,\beta)$-agnostic PAC algorithm (not necessarily stable). Secondly, we prove that $n = \Omega(d/(\gamma\alpha))$ is required even for the realizable setting. The proof of the lower bound is based on a similar bound for private prediction algorithms \citep[Theorem~2.2]{DworkFeldman18}.

One way to interpret this bound is that $\alpha$-uniform stability can be achieved essentially ``for free", that is, using asymptotically the same number of samples as is necessary for learning. In contrast, known general approaches for achieving stability do not give any non-trivial guarantees without increasing the sample complexity by at least a factor of $1/\gamma$ \citep{ShwartzSSS10}.

This uniformly stable algorithm easily implies existence of an $(\epsilon,0)$-differentially private prediction learning algorithm with sample complexity of $\widetilde{O}(d/(\epsilon\alpha^2))$. This step can be done using the aggregation technique in \citep{DworkFeldman18} but we give an even simpler conversion.

\begin{corollary}
 \label{cor:sta-to-pri}
		For every class $H$ of VC dimension $d$ and $\alpha,\beta,\eps \in (0,1)$, there exists an $(\alpha,\beta)$-agnostic PAC learning algorithm with $\epsilon$-differentially private prediction that has sample complexity of $\tilde{O}(d/\epsilon \alpha^2)$.
\end{corollary}
This bound implies that for $\eps = \Omega(1)$, $\eps$-differentially private prediction can be achieved without increasing the sample complexity (asymptotically).

One limitation of the bound in Corollary \ref{cor:sta-to-pri} is that it only improves on learning with differential privacy when $\alpha$ is  sufficiently large. Specifically, the sample complexity of $\eps$-DP $(\alpha,\beta)$-agnostic PAC learning is $\tilde O(r/(\alpha\eps) + d/\alpha^2)$, where $r$ is the representation dimension of $H$ \citep{BeimelNS:13}. Thus Corollary \ref{cor:sta-to-pri} gives a better bound than samples complexity of $\eps$-DP learning only when $\alpha \geq d/r$ (for most known classes of functions $H$, $r = \Theta(\log(|H|))$
\citep{FeldmanXiao15}). The only known result in which learning with $\eps$-private prediction requires fewer samples than $\eps$-DP learning for all $\alpha$ is the algorithm for learning thresholds on a line with sample complexity of $\tilde O(1/(\eps\alpha) + 1/\alpha^2)$  \citep{DworkFeldman18}. Note that the VC dimension of this class is $1$ while the representation dimension is $\Theta(\log |X|)$.

Our main technical results is the first general bound that for a number of classes improves on $\eps$-DP $(\alpha,\beta)$-agnostic PAC learning in all parameter regimes. Specifically, we describe an algorithm with sample complexity $\tilde{O}(d^2/(\epsilon\alpha)+d/\alpha^2)$:
	\begin{theorem} \label{thm:main}
For every class $H$ of VC dimension $d$ and $\alpha,\beta,\eps \in (0,1)$, there exists an $(\alpha,\beta)$-agnostic PAC learning algorithm with $\epsilon$-differentially private prediction that has sample complexity of
		\[
		n \le \tilde{O}\left( \frac{d^2}{\epsilon \alpha} + \frac{d}{\alpha^2}\right).
		\]
	\end{theorem}
For comparison, $\eps$-DP learning of linear threshold functions over $B^d$, where $B\subseteq \R$ has sample complexity of $\Theta(d^2\log(|B|)/(\eps\alpha)+ d/\alpha^2)$ \citep{FeldmanXiao15}. Thus the first term in our bound is better by a factor of $\log(|B|)$. Another example is the class consisting of (the indicator function of) lines on a plane over a finite field $GF(p)$. Its VC dimension is $1$ while its representation dimension is $\Theta(\log p)$ \citep{FeldmanXiao15}. Thus our bound is better by a factor of $\log p$. Although these gaps are may appear to be small, we believe that there is an important conceptual difference between bounds in terms of size of $H$ and those depending on the VC dimension, as apparent also in the classical learning theory. In particular, our bounds apply to infinite hypothesis classes.

The best known lower bound for the problem is $\Omega(d/(\epsilon\alpha) + (d + \log(1/\beta))/\alpha^2)$ \citep[Theorem~2.2]{DworkFeldman18}. Thus the bound in Theorem~\ref{thm:main} is tight with respect to $\epsilon$ and $\alpha$, however, there is a gap in the dependence on $d$. Closing this gap is an interesting open problem.

\paragraph{Overview of the techniques:}
The algorithm we use to prove Theorem \ref{thm:main-stability} combines two standard tools. First, we pick a random subset $T$ of $S$ of size $\tilde O(d/\alpha)$, or approximately a $\gamma$ fraction of $S$. Classical results in VC theory imply that $T$ is an $\alpha$-net for $H$ with high probability. We can therefore use it to define an $\alpha$-cover $H_T$ of $H$ that has size of at most $(1/\alpha)^{O(d)}$. We then use the exponential mechanism over $H_T$ as in \citep{KasiviswanathanLNRS11} to sample a hypothesis $h\sim H_T$. Finally, we predict according to $h$, outputting $h(x)$. A simple argument implies stability: firstly, the random choice of the subset $T$ ensures that each example has small influence on the choice $H_T$ and secondly, the exponential mechanism ensures that each example has small influence on the final output function. We remark that similar uses of an exponential mechanism over a data-dependent cover have appeared in prior and subsequent work on differentially private learning \citep{ChaudhuriHsu:11,BeimelNS:13approx,beimel2015learning, bassily2019limits,bassily2020private}.

The above stable algorithm can be converted to a differentially private prediction algorithm (Corollary \ref{cor:sta-to-pri}).
In order to do so, we simulate it with $\gamma = \eps\alpha$ and, in addition, randomly flip the output label with probability $\alpha$. This additional noise increases the error by at most $\alpha$. At the same time it ensures that both $0$ and $1$ are output with probability at least $\alpha$. In particular, the additive guarantee of $\eps\alpha$ implies a multiplicative guarantee of at most $(\alpha+\eps\alpha)/\alpha = 1+\eps \leq e^\eps$.

Next, we describe the second differentially private algorithm, from Theorem~\ref{thm:main}.
The first step also uses an exponential mechanism over a cover defined using a random subset of $S$. The function output by this step is used to re-label $S$, and the result is used as a training set for the private prediction algorithm from \citep{DworkFeldman18} that assumes a realizable setting. It provides the final output. The proof relies on a substantially more delicate analysis of the differential privacy of the exponential mechanism when the set of functions it is applied to changes, combined with the effect of privacy amplification by subsampling \citep{KasiviswanathanLNRS11}. Although the final step of the algorithm and some elements of the proof are borrowed from the general technique of \citep{beimel2015learning}, new techniques and ingredients are required to derive the final bound.

We remark that an alternative way to derive Corollary \ref{cor:sta-to-pri} is to use the general relabeling approach for converting a realizable PAC learning algorithm to an agnostic one \citep{beimel2015learning}. The algorithm and the analysis resulting from this approach are more involved than our proof of Corollary \ref{cor:sta-to-pri} (we use elements of this approach in the proof of Theorem \ref{thm:main}). This approach has been communicated to us by \citet{NissimStemmer18:pc} and has been the starting point for this work. The details of this technique can be found in an (independent) work of \citet{NandiB20} which applies this result to answering multiple prediction queries with privacy. 	
\printbibliography

\else
\begin{abstract}
	We study binary classification algorithms for which the prediction on any point is not too sensitive to individual examples in the dataset. Specifically, we consider the notions of uniform stability (Bousquet and Elisseeff, 2001) and prediction privacy (Dwork and Feldman, 2018). Previous work on these notions shows how they can be achieved in the standard PAC model via simple aggregation of models trained on disjoint subsets of data. Unfortunately, this approach leads to a significant overhead in terms of sample complexity. Here we demonstrate several general approaches to stable and private prediction that either eliminate or significantly reduce the overhead. Specifically, we demonstrate that for any class $C$ of VC dimension $d$ there exists a $\gamma$-uniformly stable algorithm for learning $C$ with excess error $\alpha$ using $\tilde O(d/(\alpha\gamma) + d/\alpha^2)$ samples. We also show that this bound is nearly tight. For $\eps$-differentially private prediction we give two new algorithms: one using $\tilde O(d/(\alpha^2\eps))$ samples and another one using $\tilde O(d^2/(\alpha\eps) + d/\alpha^2)$ samples. The best previously  known  bounds for these problems are $O(d/(\alpha^2\gamma))$ and $O(d/(\alpha^3\eps))$, respectively.
	\end{abstract}
	
    \section{Preliminaries} \label{sec:prelim}
	
	
	
	\paragraph{Notation:}
	We denote by $X$ the domain, and by $Y = \{0,1\}$ the label set. We use $H$ to denote the underlying hypothesis class of functions from $X$ to $Y$ and $d$ denotes the VC dimension of $H$. The dataset is denoted by $S = ((x_i,y_i))_{i=1}^n$, and the underlying distribution is denoted by $D$. Define $[n] = \{1,\dots,n\}$. For any $I \subseteq [n]$, denote $S_I = ((x_i,y_i))_{i\in I}$. Given a hypothesis $h \colon X \to Y$, denote the expected zero-one error of $h$ by $L_D(h) := \pr_{(x,y)\sim D}[h(x) \ne y]$ and the empirical error of $h$ by $L_S(h) = \frac{1}{n} \sum_{i=1}^n \mathbbm{1}_{h(x_i) \ne y_i}$. We use $C$ to denote positive universal constants.
		
	
	\paragraph{PAC learning for randomized algorithms:}
	We consider learning algorithms that receive a training set $S \in (X\times Y)^n$ and $x \in X$ and output a random prediction $\A^S(x)$, where $\hat{y} \in Y$. Given $S$, define the \emph{population} and \emph{training} losses of $\A^S$ (respectively) by
	\[
	L_D(\A^S) \dfn \pr_{(x,y)\sim D, \A}[\A^S(x) \ne y]; \quad
	L_S(\A^S) \dfn \fr{n} \sum_{i\in [n]} \pr_{\A}[\A^S(x_i) \ne y_i].
	\]
	Given $\alpha, \beta \in (0,1)$, we say that $\A$ is \emph{$(\alpha,\beta)$-agnostic PAC} learner for $H$ if for every distribution $D$ over $X\times Y$,
	\[
	\pr_{S\sim D^n}\left[L_D(\A^S) \le \inf_{h \in H} L_D(h) + \alpha\right] \ge 1-\beta.
	\]
	
	\paragraph{Uniform stability and prediction privacy:}\label{sec:prel:priv-sta}
	
	A $\sta$-\emph{uniformly stable} (or, $\gamma$-\emph{stable} for brevity) learner is a learner whose prediction probabilities at every point change by at most an additive $\sta$ when one example in $S$ is changed. Formally, for any $S, S' \in (X\times Y)^n$ which differ in at most one example and any $x \in X$ and $y \in Y$,
	\[
	\pr_{\A}[\A^S(x) = y] \le \pr_{\A}[\A^{S'}(x) = y] + \sta.
	\]
	
	The notion of differentially private prediction \citep{DworkFeldman18} is an application of the definition of differential privacy \citep{DworkMNS:06} to learning in the setting where the only output that is exposed to (potentially adversarial) users is predictions on their points. Formally, given $\epsilon,\delta \ge 0$, we say that an algorithm $\A$ gives $(\epsilon,\delta)$-private prediction if
	\[
	\forall S,S', x, y: \quad
	\pr_{\A}[\A^S(x) = y] \le e^\eps \pr_{\A}[\A^{S'}(x) = y]  + \delta.
	\]
Postprocessing guarantees of differential privacy imply that any learning algorithm that outputs a predictor with $(\eps,\delta)$-differential privacy also gives $(\eps,\delta)$-private prediction.	We say that an algorithm gives $\epsilon$-private prediction when $\delta =0$. Note that any $\epsilon$-private algorithm is $(e^\epsilon-1)$-stable and $(0,\delta)$-private prediction is exactly the same as $\delta$-stability.

\subsection{Technical preliminaries}
\label{sec:def-for-proof}
	
\label{sec:prel:pr:learn}
	
	\paragraph{Nets for VC classes:}
	Fix $\alpha \in (0,1)$, a hypothesis class $H$ and a distribution $D$ over $X\times Y$. Given a subset $A \subseteq X$, we say that $A$ is an \emph{$\alpha$-net for $H$ with respect to $D$} if it satisfies the following: any $h, h' \in H$ that satisfy $h(x) = h'(x)$ for all $x\in A$, also satisfy $\pr_{(x,y) \sim D} [h(x) \ne h'(x)] \le \alpha$. We say that $A$ is an \emph{$\alpha$-net for $H$ with respect to $S$} if it is an $\alpha$-net with respect to the uniform distribution over $S$.
	The following is a fundamental theorem in machine learning (see, e.g.~\citealp[Section~28.3]{shalev2014understanding}).
	\begin{lemma} \label{lem:eps-net-3}
		Let $D$ be a distribution over $X \times Y$, fix  $\alpha,\beta \in (0,1)$, and let $U$ be a set of $n' = O((d \log(1/\alpha) +\log(1/\beta))/\alpha)$ i.i.d. samples from $D$. Then, with probability at least $1-\beta$, $U$ is an $\alpha$-net for $H$ with respect to $D$. Furthermore, this holds also if the samples are selected \emph{without replacement} (namely, re-sampling from $D$ until one gets $n'$ distinct elements).
	\end{lemma}
	Note that this theorem is usually stated for i.i.d. samples \emph{with replacement}. However, since repetitions do not matter for the definition of $\epsilon$ nets, one can only gain from sampling \emph{without replacement}. 
	
	\paragraph{The class $H_T$ of all possible labelings of a subset $T \subseteq X\times Y$:}
	Given a subset $T \subseteq X \times Y$, we will create a hypothesis class $H_T$ in the following manner:
	define an equivalence relation $\sim_{T}$ over $H$, by $h \sim_{T} g$ if $h(x) = g(x)$ for all $(x,y) \in T$. Then, $H_T$ will contain one representative from each equivalence class (chosen arbitrarily).
	
	\paragraph{The growth function and the Sauer-Shelah lemma:}
	Given a hypothesis class $H$, the \emph{growth function of $H$} is defined as $\tau_n(H) := \sup_{A \subseteq X \colon |A| = n} |H_A|$.
	The well-known Sauer-Shelah lemma states (see, e.g., \citealp[Lemma~6.10]{shalev2014understanding}):
	\begin{lemma} \label{lem:sauer-shelah}
		For any hypothesis class $H$ of VC dimension $d$ and $n > d + 1$,
		\[
		\tau_n(H)
		\le \sum_{i=0}^{d} \binom{n}{i}
		\le (em/d)^d.
		\]
	\end{lemma}
	
	\paragraph{Uniform convergence bounds:}
	The following is a standard uniform bound on the estimation error of a hypothesis class $H$ (e.g.~\citealp[Theorem~6.8]{shalev2014understanding}):
	\begin{lemma} \label{lem:uniform-conv}
		Fix $\alpha, \beta \in (0,1)$ and assume that $n \ge \Omega((d+\log(1/\beta))/\alpha^2)$. Then,
		\[
		\pr_{S\sim D^n}\left[\forall h \in H,~ |L_D(h) - L_S(h)| \le \alpha\right] \ge 1-\beta.
		\]
	\end{lemma}
	
\label{sec:prel:pr:pri}
	
	\paragraph{The exponential mechanism:}
	The exponential mechanism is a well-known $\epsilon$-differentially private algorithm for selecting a candidate that approximately maximizes some objective that has low sensitivity \citep{McSherryTalwar:07}. Following \citep{KasiviswanathanLNRS11} we apply it to a hypothesis class $H$, dataset $S$ and privacy parameter $\eps$. Namely $\A^{\exp}_{H,\epsilon}(S)$ samples $h \in H$ with probability proportional to $\exp(-n L_S(h)\epsilon/2)$. It is $\eps$-differentially private (in the usual sense of \citep{DworkMNS:06}) and, in particular, it gives $\eps$-private predictions. The utility guarantees for the expected function can for example be found in \citep{SteinkeU17subg}.
	\begin{lemma} \label{lem:exp-weights}
		The exponential mechanism $\A^{\exp}_{H,\epsilon}$ gives $\epsilon$-private predictions. Additionally, for any $\alpha \in (0,1)$ and $S \in (X \times Y)^n$, if $n \ge 2\ln(|H|)/(\epsilon \alpha)$ then
		\[
		L_S(\A^{\exp}_{H,\epsilon}(S)) \le \inf_{h \in H} L_S(h) + \alpha .
		\]
	\end{lemma}
	
	\paragraph{Privacy amplification by subsampling:}
	
	The following lemma is an adaptation of the standard privacy amplification-by-sampling technique \citep{KasiviswanathanLNRS11} to private prediction.
	
	\begin{lemma} \label{lem:amp-by-subs}
		Let $\A'$ be an algorithm operating on a sample of size $n' = \eta n$ for $\eta \in (0,1)$ and let $\A$ be the following algorithm that receives a sample of size $n$:
		\begin{enumerate}
			\item Select a uniformly random subset $T \subseteq S$ of size $n'$.
			\item Run $\A'$ on $T$.
		\end{enumerate}
		For any $\epsilon \in (0,1)$, if $\A'$ gives $\epsilon$-private prediction then $\A$ gives $2\epsilon\eta$-private prediction.
	\end{lemma}

	\paragraph{Private prediction in the realizable setting:}
	\citet[Theorem~4.1]{DworkFeldman18} have shown that $\tilde O(d/(\epsilon\alpha))$ samples suffice to learn with private prediction in the realizable setting (that is when there exists $h\in H$ with zero error).
	\begin{lemma} \label{lem:realizable-alg}
		For any $\epsilon,\alpha,\beta$ and $H$, there exists an $(\alpha,\beta)$-PAC learner with $\eps$-private prediction $\A^R_{\epsilon,\alpha,H}$ for distributions realizable by $H$ with sample complexity of $n = \tilde O(d/(\alpha \epsilon))$.
	\end{lemma}

\section{Uniform Stability of PAC Learning} \label{sec:uni-sta}
In this section we describe the uniformly stable PAC learning algorithm, prove a (nearly) matching lower bound on its sample complexity and derive the corollary for private prediction.

For convenience, we start by restating Theorem \ref{thm:main-stability}.
\begin{theorem}[Thm.~\ref{thm:main-stability} restated]
\label{thm:main-stability-body}
For every class $H$ of VC dimension $d$ and $\alpha,\beta,\gamma \in (0,1)$, there exists a $\gamma$-uniformly stable $(\alpha,\beta)$-agnostic PAC learning algorithm with sample complexity of
		\[
		n = O\left( \frac{d \log(d/\alpha)}{\gamma \alpha}
		+ \frac{d + \log(1/\beta)}{\alpha^2}\right).
		\]\end{theorem}
We outline the algorithm and its analysis below. The details of the proof appear in Section \ref{sec:pr-sta}.
\paragraph{Proof outline:} The algorithm, that receives as input $S \in (X\times Y)^n$ and $x \in X$, consists of two steps:
\begin{itemize}
	\item Randomly select a subset $T \subseteq S$ of size $|T| = \gamma n/2$, and create the hypothesis class $H_T$, as defined in Section~\ref{sec:prel:pr:learn} ($H_T$ contains all classifications of $H$ on $T$).
	\item Run the exponential mechanism (from Section~\ref{sec:def-for-proof}) on the set of hypotheses $H_T$ evaluated on the set of examples $S$ with privacy parameter $\gamma/4$, to draw a hypothesis $\hat{h} \sim \A^{\exp}_{H_T,\gamma/4}(S)$. Output $\hat{h}(x)$.
\end{itemize}
The following properties are useful for the proof:
\begin{enumerate}
	\item With high probability,
	\begin{equation}\label{eq:113}
	\inf_{h \in H_{T}} L_S(h) \le \inf_{h \in H} L_S(h) + \alpha/3.
	\end{equation}
	This follows from the fact that $T = \widetilde{\Omega}(d/\alpha)$, hence it is an $\alpha/3$ net for $H$, with respect to the uniform distribution over $S$ (Lemma~\ref{lem:eps-net-3}).
	\item From Sauer-Shelah lemma (Lemma~\ref{lem:sauer-shelah}), the cardinality of $H_T$ is upper-bounded by $|H_T| \le O(n^d)$.
	\item The exponential mechanism in the second step is $\gamma/4$-private, hence its stability parameter is $(e^{\gamma/4}-1)\le \gamma/2$. By Lemma \ref{lem:exp-weights}, $|S| = O(\log |H_T|/(\gamma \alpha)) = \widetilde{O}(d/(\gamma\alpha))$ suffices to ensure that
	\begin{equation}\label{eq:114}
	\E_{\hat{h}\sim \A^{\exp}_{H_T,\gamma/4}(S)}[L_S(\hat{h})] \le \inf_{h \in H_{T}} L_S(h) + \alpha/3.
	\end{equation}
\end{enumerate}

The algorithm has excess error of at most $\alpha$: combining \eqref{eq:113} and \eqref{eq:114}, we get that with high probability over the choice of $T$, $\E_{\hat{h}}[L_S(\hat{h})] \le \inf_{h\in H} L_S(h) + 2\alpha/3$, and from uniform convergence (Lemma~\ref{lem:uniform-conv}) we obtain that $\E_{\hat{h}}[L_D(\hat{h})] \le \inf_{h\in H} L_D(h) + \alpha$, with high probability over the choice of $S$ and $T$. Next, we claim that it is $\gamma$-stable: assume that $S$ and $S'$ are two training sets that differ in one example. This example has a probability of at most $\gamma/2$ to appear in $T$ and affect the creation of $H_T$. Additionally, the exponential mechanism is $\gamma/2$ stable. Composing the stability bounds from both steps we obtain that the algorithm is $\gamma$-stable as desired.

The tightness of the upper bound is implied by the following lower bound.
\begin{theorem} \label{thm:sta-lb}
	Let $H$ be a class of functions of VC dimension $d$. Fix $\alpha \le 1/4$. Let $\A$ be a PAC learning algorithm that is $\gamma$-stable and for any realizable distribution $D$ over $X \times Y$,
	$\E_{S\sim D^n}[L_{D}(\A^S)] \le \alpha$.
	Then, $n \ge (d-1)/(8\gamma\alpha)$.
\end{theorem}
The proof follows the same structure as the proof of Theorem~2.2 in \citep{DworkFeldman18} and appears in Appendix~\ref{sec:pr-lb}.

\paragraph{Prediction privacy via stability:}  \label{sec:sketch:sta-implies-pri}
We now observe that any $\alpha\eps/2$-stable $(\alpha,\beta)$-PAC learning algorithm can be converted to a $(2\alpha,\beta)$-PAC learning algorithm with $\eps$-private prediction simply by flipping the predictions with probability $\alpha$.

More formally, the algorithm consists of the following two steps:
\begin{enumerate}
	\item Run the $\gamma$-stable algorithm from Theorem~\ref{thm:main-stability} with $\gamma = \epsilon\alpha/2$ and approximation parameter $\alpha$, and let $y'$ be the output prediction.
	\item With probability $1-\alpha$ predict $y'$ and otherwise predict $1-y'$.
\end{enumerate}

Let $\A$ denote the algorithm. Its error is trivially bounded by $2\alpha$. Privacy follows from definition, using the facts that $\A$ is $\epsilon\alpha/2$-stable and that for any $S \in (X\times Y)^n$, $x \in X$ and $y \in Y$, $\pr[\A(S,x) = y] \ge \alpha$. Thus we obtain Corollary \ref{cor:sta-to-pri}.

\iffull
\iffull
\subsection{Proof of Theorem~\ref{thm:main-stability}} \label{sec:pr-sta}
\else
\section{Proof of Theorem~\ref{thm:main-stability}} \label{sec:pr-sta}
\fi

The algorithm receives as an input a training set $S$ and a point $x \in X$, and outputs $y \in \{0,1\}$ as a prediction for $x$ defined below:
\begin{enumerate}
	\item Select a uniformly random subset $I \subseteq [n]$ of size $n'$, where $n' \in [n]$ is a parameter to be defined later ($n'$ should be thought of as $n' \approx \gamma n$).
	\item Create the hypothesis class $H_{S_I}$ containing all possible labelings of $S_I$, as defined in Section~\ref{sec:prel:pr:learn}.
	\item Execute the exponential mechanism $\A_{H_{S_I},\gamma}^{\exp}(S)$ with privacy parameter $\gamma$ on the sample $S$ to randomly select a hypothesis $h_{S,S_I} \in H_{S_I}$, and output $h_{S,S_I}(x)$.
\end{enumerate}

We proceed with the formal definition. First, for any $S \in (X\times Y)^n$ and $T \in (X\times Y)^{n'}$, define $h_{S,T} = \A_{H_{T},\gamma}^{\exp}(S)$, viewing $h_{S,T}$ as a function from $X \to [0,1]$. Note that we do not require $T \subseteq S$ in this definition. Next, define
\begin{equation*}
\A^{\mathrm{sta}}(S) := \frac{1}{\binom{n}{n'}} \sum_{I \subseteq [n] \colon |I| = n'} h_{S, S_I}.
\end{equation*}
The final prediction of the algorithm given $x \in X$ is $\A^{\mathrm{sta}}(S)(x)$.

We are ready to state the main lemma:
\begin{lemma} \label{lem:stable-main}
Let $N^{\mathrm{net}}_{\alpha,\beta,d}$ denote the smallest number $n'$ that suffices for Lemma~\ref{lem:eps-net-3} to hold, given parameters $\alpha,\beta$ and $d$ (the minimal size required for a random set to be an $\alpha$-net with probability $1-\beta$). Let $N^{\exp}_{k,\epsilon,\alpha}$ denote the sample complexity required for $\epsilon$-differentially private exponential mechanism to be $\alpha$-approximate given a hypothesis class of size $k$.
	Assume that
	\[
	n' \ge N^{\mathrm{net}}_{\alpha, \alpha, d}; \quad
	n \ge N^{\exp}_{\tau_{n'}(H),\gamma,\alpha}; \quad
	n \ge n'/\gamma.
	\]
	Then
	\[
	L_S(\A^{\mathrm{sta}}(S)) - \inf_{h \in H} L_S(h) \le 3\alpha
	\]
	and $\A^{\mathrm{sta}}$ is $3\gamma$-stable.
\end{lemma}
First, we prove Theorem~\ref{thm:main-stability} using Lemma~\ref{lem:stable-main}, and then we prove this lemma.
\begin{proof}[Proof of Theorem~\ref{thm:main-stability}]
	Given $\alpha, \beta$ and $\gamma$, we will show an algorithm with the desired sample complexity.
	It suffices to find an algorithm which is $O(\gamma)$-stable and $(O(\alpha),O(\beta))$-agnostic PAC.
	To do so, we apply the above algorithm $\A^{\mathrm{sta}}$ with parameter $n' = N^{\mathrm{net}}_{\alpha,\alpha,d}$, and then select $n$ as the smallest integer which satisfies the conditions of Lemma~\ref{lem:stable-main}.
	
	We apply Lemma~\ref{lem:stable-main} to get a bound on the empirical error of $\A^{\sta}(S)$ and then apply uniform convergence to generalize to the data distribution. In particular, from Lemma~\ref{lem:uniform-conv}, with probability $1-\beta$ over $S$,
	\[
	L_D(\A^{\mathrm{sta}}(S))
	\le L_S(\A^{\mathrm{sta}}(S)) + \alpha
	\le \inf_{h \in H} L_S(h) + 4\alpha
	\le \inf_{h \in H} L_D(h) + 5\alpha.
	\]
\end{proof}
Lastly, we prove Lemma~\ref{lem:stable-main}:
\begin{proof}[Proof of Lemma~\ref{lem:stable-main}]
First we prove $3\alpha$-approximation. From the condition that $n \ge N^{\exp}_{\tau_{n'}(h),\epsilon,\alpha}$, for any $I$,
	\begin{equation*}
	L_S(h_{S,S_I})
	\le \min_{h \in H_{S_I}} L_S(h) + \alpha.
	\end{equation*}
	Since $n' \ge N^{\mathrm{net}}_{\alpha, \alpha, d}$, with probability at least $1-\alpha$ over the choice of $I$:
	\begin{equation*}
	\min_{h \in H_{S_I}} L_S(h) \le \inf_{h \in H} L_S(h) + \alpha.
	\end{equation*}
	Thus,
	\[
	L_S(\A^{\mathrm{sta}}(S))
	= \frac{1}{\binom{n}{n'}} \sum_{I \subseteq [n] \colon |I| = n'} L_S(h_{S, S_I})
	\le \alpha + \frac{1}{\binom{n}{n'}} \sum_{I \subseteq [n] \colon |I| = n'} \min_{h \in H_{S_I}} L_S(h)
	\le \inf_{h \in H} L_S(h) + 3\alpha.
	\]

Now we prove the $3\gamma$-stability.
	Let $S$ be a sample and let $S'$ be obtained from $S$ by replacing one sample. Without loss of generality we can assume that $S'$ is obtained by removing $(x_1,y_1)$ from $S$ and adding $(x_1',y_1')$. Our goal is to show that $\A^{\mathrm{sta}}(S) \preceq \A^{\mathrm{sta}}(S') + 3\gamma$.
	
	For any $T$, Lemma~\ref{lem:exp-weights} implies that
	\[
	h_{S,T}
	\preceq e^{\gamma} h_{S',T}
	\preceq (1+2\gamma) h_{S',T}
	\preceq h_{S',T} + 2\gamma,
	\]
	using the inequality $e^x \le 1+2x$ for $x \le 1$.
	To conclude the proof, using the fact that $n'/n \le \gamma$,
	\begin{align*}
	\A^{\mathrm{sta}}(S)
	&= \frac{1}{\binom{n}{n'}} \sum_{I \subseteq [n] \colon |I| = n'} h_{S, S_I}
	\preceq \frac{1}{\binom{n}{n'}} \sum_{I \subseteq [n] \colon |I| = n'} h_{S', S_I} + 2 \gamma \\
	&= \frac{1}{\binom{n}{n'}} \sum_{I \subseteq [n] \setminus \{1\} \colon |I| = n'} h_{S', S_I}
	+ \sum_{I \subseteq [n] \colon |I| = n', 1 \in I} h_{S', S_I} +2\gamma \\
	&\preceq \A^{\mathrm{sta}}(S') + \frac{n'}{n} + 2\gamma
	\preceq \A^{\mathrm{sta}}(S') + 3\gamma.
	\end{align*}
\end{proof}	

\begin{remark} \label{rem:1}
	It is possible to slightly improve the bound in Theorem~\ref{thm:main-stability}, replacing $\log(d/\alpha)$ with $\log(1/\alpha)$. In order to do so, one has to replace $H_{S_I}$ with an $\alpha$-net, namely, a set $G_{S_I}$ of hypotheses from $H$ which satisfies:
	\[
	\forall h \in H, \exists g \in G_{S_I}, \text{ s.t. }
	| \{ (x,y) \in S_I \colon h(x) \ne g(x) \}| \le \alpha |I|.
	\]
	There exists such a set of cardinality $(1/\alpha)^{O(d)}$. Using this set, one can relax the requirement $n \ge N^{\exp}_{\tau_{n'}(H),\gamma,\alpha}$ from Lemma~\ref{lem:stable-main} to $n \ge N^{\exp}_{1/\alpha^{O(d)},\gamma,\alpha}$, and the improved bound would follow.
\end{remark}

\section{PAC Learning with Prediction Privacy}
In this section we describe our main technical result for private prediction. Our algorithm is the first general algorithm that can PAC learn an arbitrary class $H$ of VC dimension $d$ with the (nearly) optimal dependence of the sample complexity on $\eps$ and $\alpha$. However the dependence on the dimension $d$ is quadratic. More formally, we prove the following upper bound on the sample complexity of the problem.

	\begin{theorem}[Theorem \ref{thm:main} restated] \label{thm:main-body}
For every class $H$ of VC dimension $d$ and $\alpha,\beta,\eps \in (0,1)$, there exists an $(\alpha,\beta)$-agnostic PAC learning algorithm with $\epsilon$-differentially private prediction that has sample complexity of
		\[
		n \le \tilde{O}\left( \frac{d^2}{\epsilon \alpha} + \frac{d}{\alpha^2}\right).
		\]
	\end{theorem}

We present the algorithm and the outline of its analysis below. Additional technical details are given in Section~\ref{sec:pr:pri-alg}.

\subsection{Overview of the algorithm and its analysis}
\label{sec:main-outline}
The algorithm that achieves the claimed sample complexity is described below (with slight simplifications for the clarity of exposition):
\begin{itemize}
	\item
	The first steps are similar to those of the stable algorithm: we draw a random subset $T \subseteq S$ and then select, using the exponential mechanism, a hypothesis $\hat{h}\in H_T$. The only difference is that $T$ is now selected to be of size $|T| = O(\epsilon |S|)$ (rather than $O(\gamma |S|)$) and the privacy parameter of the exponential mechanism is now set to be $\min(\epsilon/2, d/(\alpha |S|))$ (rather than $\gamma/4$).
	\item In the second step, we ``privatize" $\hat{h}$ as follows:
	\begin{enumerate}
		\item Let $S_{\hat{h}}$ denote the set of examples obtained by labeling the points in $S$ with $\hat{h}$, namely, $S_{\hat{h}} := \{ (x,\hat{h}(x)) \colon (x,y) \in S \}$.
		\item 
		Let $\varphi(\hat{h})$ denote the PAC learner from Lemma~\ref{lem:realizable-alg}, with sample $S_{\hat{h}}$, $d/(\alpha|S|)$ prediction privacy and excess error $O(\alpha)$.
		The final prediction is drawn according to the distribution of $\varphi(\hat{h})(x)$.
	\end{enumerate}
\end{itemize}


Let $\A$ denote the algorithm and we first establish that it is $(\alpha,\beta)$-PAC. 
As in the analysis of the stable algorithm, $\E[L_D(\hat{h})] \le \inf_{h \in H} L_D(h) + O(\alpha)$. By the guarantees of the learner from Lemma~\ref{lem:realizable-alg},
$
\Pr_{(x,y)\sim D, \varphi}[\varphi(\hat{h})(x) \ne \hat{h}(x)] \le O(\alpha).
$
Combining the above two inequalities:
\[
\E[L_D(\A^S)]
= \E[L_D(\phi(\hat{h}))] \le L_D(\hat{h}) +O(\alpha) \le \inf_{h \in H} L_D(h) + O(\alpha).
\]



Next, we explain why the algorithm gives $\epsilon$-private prediction. Denote by $h_{S,T}(x)$ the random prediction given a training set $S$ and a fixed subset $T$: $h_{S,T}(x)$ is obtained by first selecting $\hat{h}$ from $H_T$ using the exponential mechanism and then outputting a random value drawn according to the distribution of $\varphi(\hat{h})(x)$. For the analysis, we extend the definition of $h_{S,T}(x)$ to sets $T$ which are not necessarily subsets of $S$.
Our goal is to show that for any $x \in X$ and $y \in Y$,
\begin{equation}\label{eq:21}
\pr_{T \subseteq S, h_{S,T}(x)}[h_{S,T}(x) = y] \le e^\epsilon \pr_{T' \subseteq S', h_{S',T'}(x)}[h_{S',T'}(x) = y],
\end{equation}
where the probability is both over the selections of the subsets $T$ and $T'$ and over the random selections of $h_{S,T}(x)$ and $h_{S',T'}(x)$. We split \eqref{eq:21} in two, first comparing $h_{S,T}$ with $h_{S,T'}$ and then $h_{S,T'}$ with $h_{S',T'}$: for any $x \in X$ and $y \in Y$,
\begin{align}
&\pr_{T \subseteq S,h_{S,T}(x)}[h_{S,T}(x) = y]
\le e^{\epsilon/2} \pr_{T' \subseteq S', h_{S,T'}(x)}[h_{S,T'}(x) = y]; \ \text{and} \label{eq:116}\\
&\pr_{T' \subseteq S',h_{S,T'}(x)}[h_{S,T'}(x) = y] \le e^{\epsilon/2} \pr_{T' \subseteq S',h_{S',T'}(x)}[h_{S',T'}(x) = y] \label{eq:115}.
\end{align}
\eqref{eq:115} follows from the fact that for any fixed $T'$, $h_{S,T'}(x)$ is the composition of two private algorithms: first, the exponential mechanism selects $\hat{h}$ privately, and then $\varphi(\hat{h})(x)$ is a private prediction given any fixed $\hat{h}$.

Next, we sketch the proof of \eqref{eq:116}.
The terms in both sides of the inequality can be calculated as follows: first $T$ and $T'$ are randomly drawn as subsets of $S$ and $S'$, respectively. Then, the random predictions $h_{S,T}(x)$ and $h_{S,T'}(x)$ are made. Due to this structure, we can use privacy \emph{amplification by sub-sampling} (Lemma~\ref{lem:amp-by-subs})\footnote{To prove \eqref{eq:116}, one cannot use Lemma~\ref{lem:amp-by-subs} as stated. However, the proof in our case is analogous.} to compare $h_{S,T}(x)$ with $h_{S,T'}(x)$. In particular, we prove that $h_{S,T}(x)$ is $O(1)$-private as a function of $T$, and since $|T| = O(\epsilon |S|)$, the sub-sampling boosts the privacy by a factor of $O(\epsilon)$ and \eqref{eq:116} follows. Hence, we are left with proving that $h_{S,T}(x)$ is $O(1)$-private as a function of $T$:
\begin{equation} \label{eq:117}
\pr_{h_{S,T}(x)}[h_{S,T}(x) = y] \le e^{O(1)} \pr_{h_{S,T'}(x)}[h_{S,T'}(x) = y],
\end{equation}
for (almost) any $T$ and $T'$ that differ on one element.

To prove \eqref{eq:117}, we fix $T$ and $T'$ that differ in one element and create a matching between the elements of $H_T$ and the elements of $H_{T'}$. This matching is either one-to-one, one-to-two or two-to-one and for any matched pair $h \in H_T$ and $h'\in H_{T'}$, the following properties hold:
\begin{itemize}
	\item $|L_{S}(h) - L_{S}(h')|\le \alpha/d$.
	\item For any $x \in X$ and $y \in Y$, $\pr[\phi(h)(x) = y] \le e^{O(1)} \pr[\phi(h')(x) = y]$, where the probability is over the random selections of $\varphi(h)(x)$ and $\varphi(h')(x)$.
\end{itemize}
These two properties imply that $\pr[h_{S,T}(x)=y]$ is within constant factors of $\pr[h_{S,T'}(x)=y]$: the first property ensures that the probability to select $h$ from $H_T$ via the exponential mechanism is within a constant factor from the probability to select $h'$ from $H_{T'}$ (for any matched $h$ and $h'$). The second property states that $\pr[\varphi(h)(x)=y]$ and $\pr[\varphi(h')(x)=y]$ are within a constant factor from each other. Hence,  \eqref{eq:117} follows from these two properties, and it remains to describe why they hold.
\begin{itemize}
	\item First, we define the aforementioned matching between $H_T$ and $H_{T'}$: $h \in H_T$ is matched with $h' \in H_{T'}$ if and only if $h(x) = h'(x)$ for all $(x,y) \in T \cap T'$.
	\item We apply Lemma~\ref{lem:eps-net-3} on $\eta$-nets, substituting $U = T \cap T'$ and $\eta = \alpha/d$. We obtain that $T\cap T'$ is an $\alpha/d$-net for $H$ with respect to the uniform distribution over $S$ (with high probability over a random selection of $T$). Any pair of matched hypotheses satisfy $h(x) = h'(x)$ for all $(x,y) \in T\cap T'$, hence, by definition of $\eta$-nets, they satisfy
	\begin{equation} \label{eq:15}
	|\{(x,y) \in S \colon h(x) \ne h'(x) \}| \le |S|\alpha/d.
	\end{equation}
	\item The first desired property, namely $|L_{S}(h) - L_{S}(h')|\le \alpha/d$, follows immediately from \eqref{eq:15}.
	\item For the second property, recall that $\phi(h)(x)$ is the output of a learner with $d/(\alpha|S|)$-private prediction, trained on $\{(x,h(x)) \colon (x,y) \in S\}$. \eqref{eq:15} implies that the training sets used to train $\varphi(h)(x)$ and $\varphi(h')(x)$ differ on at most $\alpha|S|/d$ examples. Applying the privacy guarantee $\alpha|S|/d$ times (referred to as {\em group privacy}), one obtains that $\pr[\phi(h)(x) = y] \le (e^{d/(\alpha|S|)})^{\alpha|S|/d}\pr[\phi(h')(x)=y] = e \pr[\phi(h')(x)=y]$, as required. The second property holds, and the proof follows.
\end{itemize}

\iffull
\subsection{Detailed proof} \label{sec:pr:pri-alg}
\else
\section{Proof of Theorem~\ref{thm:main}} \label{sec:pr:pri-alg}
\fi
We will start by describing an algorithm which gives $(O(\epsilon),O(\epsilon \alpha))$-private prediction (namely, $(\epsilon,\delta)$-privacy with $\delta = O(\epsilon \alpha)$, see Section~\ref{sec:prelim}). We then convert it to an algorithm with $O(\epsilon)$-private prediction as in the proof of Corollary \ref{cor:sta-to-pri}. 

The algorithm receives a training set $S$ and a point $x \in X$, and outputs a label $y$ for $x$. It is defined as follows:
\begin{enumerate}
	\item Select a subset $I \subseteq [n]$ of size $|I|=n'$ uniformly at random ($n'$ is a parameter to be defined later, and should be thought of as $n' \approx \epsilon n$).
	\item Create the hypothesis class $H_{S_I}$, as defined in Section~\ref{sec:prel:pr:learn} ($H_{S_I}$ contains all the different prediction patterns of hypotheses from $H$ on $S_I$).
	\item Draw a random $h \in H_{S_I}$ using the exponential mechanism with loss evaluated on $S$ and privacy parameter $2/(\eta n)$, where $\eta > 0$ is a parameter to be defined later (should be thought of as $\epsilon/d$).
	\item Define the dataset $S_h$ as the predictions of $h$ on $S$, $S_h := ((x_i, h(x_i)))_{i=1}^n$ (where $(x_i,y_i)$ is the $i$'th sample from $H$). Recall that $\A_{\epsilon,\alpha,H}^R$ denotes the realizable private learner from Lemma~\ref{lem:realizable-alg} with privacy $\epsilon$ and approximation error $\alpha$ (we will ensure that the sample size is sufficiently large to achieve these guarantees). Let
	$\varphi_S(h) := \A_{1/(\eta n),\alpha,H}^R(S_h)$ denote the learned hypothesis given the sample $S_h$.
	\item Output $\varphi_S(h)(x)$.
\end{enumerate}
Viewed as a deterministic function the hypothesis output by this algorithm is described as follows. First, for any $S \in (X\times Y)^n$ and $T \in (X\times Y)^{n'}$ (not necessarily $T \subseteq S$), define:
\begin{equation} \label{eq:154}
h_{S,T}
= \frac{1}{Z} \sum_{h \in H_{T}} \lambda_h \varphi_S(h), \quad \text{ where }
\lambda_h = \exp\left( -L_S(h) /\eta \right)\quad
\text{and} \quad Z = \sum_{h \in H_{S_I}} \lambda_h.
\end{equation}
Then, for any $S$, define
\begin{equation} \label{eq:2}
h_S := \frac{1}{\binom{n}{n'}} \sum_{I \subseteq [n] \colon |I| = n'} h_{S, S_I}.
\end{equation}

We now summarize the conditions that need to be satisfied to prove the main result.
\begin{lemma} \label{lem:main}
	For $\alpha,\beta,\epsilon \in (0,1)$ and a positive integer $d$, define the following values:
	\begin{itemize}
		\item Let $N^{\mathrm{net}}_{\alpha,\beta,d}$ be the smallest sample size such that for any distribution $D$ and any hypothesis class $H$ of VC dimension $d$, a random i.i.d. sample of size $N^{\mathrm{net}}_{\alpha,\beta,d}$ is an $\alpha$-net for $H$ with probability $1-\beta$.  Lemma~\ref{lem:eps-net-3} gives that $N^{\mathrm{net}}_{\alpha,\beta,d} = O((d \log(1/\alpha) +\log(1/\beta))/\alpha)$.
		\item
		Let $N^R_{\epsilon, \alpha,\beta,d}$ denote the smallest sample size such that with probability $1-\beta$ over the choice of $S$:
		\begin{equation} \label{eq:222}
		\forall h \in H, \E_{(x,y) \sim D}[|h(x) - \varphi_S(h)(x)|]\le \alpha,
		\end{equation}
		where $\varphi_S(h)$ is obtained from $h$ by applying the PAC learner with private prediction for the realizable case. As we explain in Remark~\ref{rem:priv-real}, $N^R_{\epsilon, \alpha,\beta,d}$ is essentially the same as the sample complexity stated in Lemma \ref{lem:realizable-alg}, and equals $\tilde{O}(d/(\epsilon\alpha))$.
		\item Let $N^{\exp}_{\epsilon, \alpha,k}$ denote the sample size required for the $\epsilon$-differentially private exponential mechanism to be $\alpha$-approximate for any hypothesis class of size $k$. Lemma~\ref{lem:exp-weights} implies that $N^{\exp}_{\epsilon, \alpha,k} \le 2 \ln(k)/(\epsilon \alpha)$.
		\item Let $N^G_{\alpha,\beta,d}$ denote the sample size required to ensure that for any hypothesis class $H$ of VC dimension $d$, with probability $1-\beta$ over the choice of the training set $S$:
		\[
		\forall h \in H,\ |L_D(h) - L_S(h)| \le \alpha.
		\]
		Standard uniform convergence bounds (Lemma \ref{lem:uniform-conv}) give $N^G_{\alpha,\beta,d} = \Theta((d + \log(1/\beta))/\alpha^2)$.
	\end{itemize}
	Recall that $\tau_n(H)$ is the growth function of $H$, defined in Section~\ref{sec:prel:pr:learn}.
	Let $n'$ and $\eta$ be parameters such that the following holds:
	\begin{equation*}
	n \ge N^G_{\alpha,\beta,d}; \ \
	n' \ge N^{\mathrm{net}}_{\alpha, \alpha, d}; \ \ 
	n' \ge N^{\mathrm{net}}_{\eta, \alpha, d} + 1; \ \ 
	n \ge N^{\exp}_{2/(\eta n), \alpha, \tau_n(H)}; \ \ 
	n \ge N^R_{1/(\eta n), \alpha, d}; \ \ 
	n' \le \epsilon n; \ \ 
	\epsilon \ge 1/(\eta n).
	\end{equation*}
	Then, the hypothesis $h_S$ defined in \eqref{eq:2} above gives $(O(\epsilon), O(\epsilon\alpha))$-private predictions. Additionally, for any set $S$,
	\[
	L_D(h_S) \le \inf_{h \in H} L_D(h) + O(\alpha),
	\]
	with probability $1-\beta$.
\end{lemma}

\begin{remark} \label{rem:priv-real}
	We briefly explain why $N^R_{\epsilon, \alpha,\beta,d} = \tilde{O}(d/(\epsilon\alpha))$. In Theorem~4.1 in \citep{DworkFeldman18} shows that this sample complexity is sufficient for the existence of a PAC learning algorithm with $\eps$-DP prediction for the realizable setting. This implies that for any
	\emph{fixed} $h$, with probability $1-\beta$ over the choice of $S$ labeled by $h$,
	$\E_{(x,y) \sim D}[|h(x) - \varphi_S(h)(x)|]\le \alpha$. In our case, we apply the algorithm to $S_h$, where $h$ depends on $S$. Therefore we need the learning result to hold for all possible ways to label $S$ by $h\in H$. Namely, that with probability $\ge 1-\beta$, for all $h \in H$, $\E_{(x,y) \sim D}[|h(x) - \varphi_S(h)(x)|]\le \alpha$. 
	
	The proof in \citep{DworkFeldman18} relies on having a (non-private) $(\alpha,\beta)$-PAC learner with sample complexity $O((d+\log(1/\beta))/\alpha^2)$. 
	In order for the stronger statement in \eqref{eq:222} to hold, all we need is a non-private $(\alpha,\beta)$-PAC learner that succeeds for all $h \in H$ \emph{at the same time}. Namely, that with probability $\ge 1-\beta$ over a random sample $S$ of size $\widetilde{\Theta}(d/\alpha)$ for all $h \in H$ the algorithm, given $S_h$ outputs $g$ that satisfies $\Pr_{x \sim D} [h(x) \ne g(x)] \le \alpha$. For classes of VC dimension $d$, uniform convergence results imply that any learning algorithm that outputs $g$ that is consistent with $S_h$ will have this property for sample size $m = O((d+\log(1/\beta))/\alpha^2)$. This is the same bound as the one used in \citep{DworkFeldman18} and thus we get $N^R_{\epsilon, \alpha,\beta,d} = \tilde{O}(d/(\epsilon\alpha))$.
\end{remark}

Theorem~\ref{thm:main} follows from Lemma~\ref{lem:main} by setting $n =\tilde{\Theta}(d^2/(\epsilon\alpha) + d/\alpha^2)$, $n' = \tilde{\Theta}(d^2/\alpha)$, $\eta = \tilde\Theta(\alpha/d)$. To convert the prediction from $(\epsilon,\epsilon\alpha)$-private to $\epsilon$-private we flip the value of the final output $h_S$ with probability $\alpha$ (as in the proof of Corollary \ref{cor:sta-to-pri}).
The conditions in Lemma~\ref{lem:main} have the following roles:
\begin{itemize}
	\item $n \ge N^G_{\alpha,\beta,d}$ is required for generalization (guaranteeing that the error relative to $D$ is within $\alpha$ of the error on $S$).
	\item $n' \ge N^{\mathrm{net}}_{\alpha, \alpha, d}$ is required to ensure that the sub-sampled set $T$ is an $\alpha$-net for $H$ with probability $1-\alpha$. This ensures that the hypothesis class $H_T$ contains an $\alpha$-optimal hypothesis.
	\item $n' \ge N^{\mathrm{net}}_{\eta, \alpha, d} + 1$ guarantees that $T \cap T'$ is an $\eta \approx \alpha/d$-net.
	\item
	$n \ge N^{\exp}_{2/(\eta n), \alpha, \tau_n(H)}$ guarantees that the executed exponential mechanism is $2/(\eta n)$-differentially private and $\alpha$-approximate. The condition $\epsilon \ge 1/(\eta n)$ guarantees that the exponential mechanism is $O(\epsilon)$-differentially private.
	\item $n \ge N^R_{1/(\eta n), \alpha, \beta, d}$ guarantees that the PAC learning algorithm for the realizable case is $1/(\eta n)$-private and $(\alpha,\beta)$-approximate.
	\item $n' \le \epsilon n$ is required for privacy amplification by sub-sampling from $O(1)$-privacy to $O(\eps)$-privacy.
\end{itemize}

We now prove that the algorithm we defined is an $(O(\alpha),O(\beta))$-PAC learning algorithm.
\begin{lemma} \label{lem:accurate}
	Under the conditions of Lemma~\ref{lem:main} it holds that $L_D(h_S) \le \inf_{h \in H} L_D(h) + 4\alpha$, with probability at least $1-\beta$.
\end{lemma}
\begin{proof}
Assume that $I$ is a uniformly random subset of $[n]$ of size $n'$.
	Since $n' \ge N^{\mathrm{net}}_{\alpha, \alpha, d}$, with probability at least $1-\alpha$ over the choice of $I$:
	\begin{equation*}
	\min_{h \in H_{S_I}} L_S(h) \le \inf_{h \in H} L_S(h) + \alpha.
	\end{equation*}
	In particular,
	\begin{equation*}
	\E_I \left[ \min_{h \in H_{S_I}} L_S(h) \right] \le \inf_{h \in H} L_S(h) + 2\alpha,
	\end{equation*}
	where $I$ is a uniformly random subset.

	Next, let $h'_{S,S_I}$ denote the output of the exponential mechanism on $H_{S_I}$, namely
	\[
	h'_{S,S_I} = \frac{1}{Z} \sum_{h \in H_{S_I}} \lambda_h h.
	\]
	Since the exponential mechanism is defined as $\alpha$-approximate, it holds that $L_S(h'_{S,S_I}) \le \min_{h\in H_{S_I}} L_S(h) + \alpha$.

	Recall the assumption that $n = \Omega((d+\log(1/\beta))/\alpha^2)$, or, equivalently, $n \ge N^G_{\alpha,\beta,d}$. This, combining with the previous two steps, imply that with probability $1-\beta$:
	\begin{equation} \label{eq:23}
	\E_I[L_D(h'_{S,S_I})]
	\le \E_I[L_S(h'_{S,S_I})] + \alpha
	\le \E_I \left[\min_{h\in H_{S_I}} L_S(h) \right]+ 2\alpha
	\le \inf_{h \in H} L_S(h) + 4\alpha
	\le \inf_{h \in H} L_D(h) + 5\alpha.
	\end{equation}

To go from $h'_{S,S_I}$ to $h_{S,S_I}$, note that $h_{S,S_I}= \phi_S(h'_{S,S_I})$. 
	From $n \ge N^R_{\epsilon, \alpha,\beta,d}$ it follows that with probability $\ge 1-\beta$:
	$L_D(\varphi_S(h)) \le L_D(h) + \alpha$ for all $h \in H$. In particular,
	$L_D(h_{S,S_I}) \le L_D(h'_{S,S_I})+ \alpha$. This and \eqref{eq:23} imply that with probability $\ge 1-2\beta$,
	\[
	L_D(h_S)
	= \E_I[L_D(h_{S,S_I})]
	\le \E_I[L_D(h'_{S,S_I})]+\alpha
	\le \inf_{h \in H} L_D(h) + 6 \alpha.
	\]
\end{proof}

The rest of this section is dedicated to proving that the algorithm is $(O(\epsilon),O(\epsilon\alpha))$-private.
Fix $S$ and let $S'$ be a set which is obtained from $S$ by replacing one sample. Without loss of generality, we can assume that $S'$ is obtained from $S$ by replacing $(x_1,y_1)$ with $(x_1',y_1')$. Our goal is to show that
\[
h_S = \frac{1}{\binom{n}{n'}} \sum_{|I| = n'} h_{S,S_I} \preceq e^{O(\epsilon)} h_{S'} + O(\epsilon \alpha).
\]

We will divide the summation over subsets $|I|=n'$ into three terms. Let $\mathcal{I}$ denote the set of all such subsets, and we will partition it to $\I'$, $\I''$ and $\I'''$ as follows:
\begin{itemize}
	\item $\I'$: all subsets satisfying $1 \notin I$, namely, those that do not correspond to the changed point $(x_1,y_1)$.
	\item $\I''$: those that satisfy $1 \in I$, and that samples with indices in $I\setminus \{1\}$ are \emph{not} an $\eta$-net for $H$ with respect to the uniform distribution over $S_{\{2,\dots,n\}}$. There are not many such sets: from the requirements $|I\setminus\{1\}| = n' - 1 \ge N^{\mathrm{net}}_{\eta,\alpha, d}$ and $n' \le \epsilon n$, it follows that $|\I''| \le \alpha |\I \setminus \I'| = \alpha |\mathcal{I}|n'/n \le \alpha \epsilon |\I|$.
	\item $\I'''$: the remaining sets. With respect to these, we will be able to perform the matching discussed in the proof outline.
\end{itemize}

Next, we will show prediction privacy of $h_{S,S_I}$ with respect to $I$, for $I \in \I'''$:

\begin{lemma} \label{lem:swap}
	Let $I \in \I'''$. Fix $i \in [n] \setminus I$, and let $I' = (I \setminus \{1\}) \cup \{i\}$.
	Then,
	\begin{equation} \label{eq:153}
	h_{S, S_I}
	\preceq 4e^6 h_{S, S_{I'}}
	= 4e^6 h_{S, S'_{I'}}.
	\end{equation}
\end{lemma}

\begin{proof}
	Note that the equality in \eqref{eq:153} follows from the fact that $1 \notin I'$ and that $S$ and $S'$ differ only on the first sample. Hence, we prove the inequality.
	We will create a matching between $H_{S_I}$ and $H_{S_{I'}}$ as follows:
	for any $h \in H_{S_I}$, let $P_h$ denote the set of all hypotheses $g \in H_{S_{I'}}$ that satisfy $h(x_j) = g(x_j)$ for all $j \in I\setminus\{1\}$. Intuitively, $P_h$ is the set of hypotheses matched with $h$, and it holds that $1 \le |P_h| \le 2$.
	
	We will fix $h \in H_{S_I}$ and $g \in P_h$. Since $I \in \I'''$, $I\setminus\{1\}$ is an $\eta$-net with respect to $S_{\{2,\dots,n\}}$. From definition of $\eta$ nets (Section~\ref{sec:prel:pr:learn}), since $h$ and $g$ agree on $I\setminus \{1\}$, there are at most $\eta(n-1)$ indices $j \in [n]\setminus\{1\}$ for which $h(x_j) \ne g(x_j)$. Hence, the number of samples $(x,y) \in S$ for which $h(x) \ne g(x)$ is at most $\eta (n-1) + 1 \le 2\eta n$ (we used that $1 \le 1/\epsilon \le \eta n$).
	This implies the following properties:
	\begin{itemize}
		\item It immediately follows that $|L_{S}(h) - L_{S}(g)| \leq 2\eta$.
		\item Let $\lambda_h$ and $\lambda_g$ denote the coefficients of $h$ and $g$ in the exponential weight mechanism corresponding to the definitions of $h_{S,S_I}$ and $h_{S,S_{I'}}$, respectively (as defined in \eqref{eq:154}). We will bound $\lambda_h$ in terms of $\lambda_g$:
		\[
		\lambda_h
		= e^{-L_{S}(h)/\eta}
		\le e^{-L_{S}(g)/\eta + 2}
		= \lambda_g e^2.
		\]
		\item Let $Z_I$ and $Z_{I'}$ denote the values of $Z$ in the definitions of $h_{S,S_I}$ and $h_{S,S_{I'}}$, respectively. It holds that
		\[
		Z_I = \sum_{h \in H_{S_I}} \lambda_h
		\le e^2 \sum_{h\in H_{S_I}} \sum_{g \in P_h} \lambda_g
		\le 2 e^2 \sum_{g \in H_{S_{I'}}} \lambda_g
		= 2 e^2 Z_{I'},
		\]
		using the fact that for any $g \in H_{S_{I'}}$, there exist at most two hypotheses $h\in H_{S_I}$ for which $g \in P_h$. Similarly, one can derive $Z_{I'} \le 2e^2 Z_I$.
		\item
		We will compare $\varphi_{S}(h)$ with $\varphi_{S}(g)$. Recall that $\varphi_{S}(h)$ is the output of a $1/(\eta n)$-private realizable learner on the training set $S_h$ (and similarly for $\varphi_{S}(g)$ and $S_g$). Since $h$ and $g$ predict differently on at most $2\eta n$ samples from $S$, $S_h$ and $S_g$ differ on at most $2\eta n$ entries, hence $\varphi_{S}(h) \preceq (e^{1/(\eta n)})^{2\eta n} \varphi_{S}(g)= e^2 \varphi_S(g)$.
	\end{itemize}
	
	Combining the above inequalities, one obtains that
	\[
	h_{S,S_I}
	= \frac{1}{Z_I} \sum_{h \in H_{S_I}} \lambda_h \varphi_{S}(h)
	\preceq \frac{2 e^6}{Z_{I'}} \sum_{h \in H_{S_I}} \sum_{g \in P_h} \lambda_g \varphi_{S}(g)
	\preceq \frac{4 e^6}{Z_{I'}} \sum_{g \in H_{S_{I'}}} \lambda_g \varphi_{S}(g)
	= 4 e^6 h_{S,S_{I'}}.
	\]
\end{proof}

Using an adaptation of \emph{amplification by sub-sampling} (discussed in Section~\ref{sec:prel:pr:learn}) and using Lemma~\ref{lem:swap}, one can bound the summation over $I \in \I'''$:
\[
\sum_{I\in \I'''} \frac{1}{\binom{n}{n'}} h_{S,S_I}
\preceq
\sum_{I\in \I'''} \sum_{i \in [n] \setminus I} \frac{4 e^6}{(n-n')\binom{n}{n'}}
h_{S, S'_{(I \setminus \{1\})\cup \{i\}}}
\preceq
\sum_{J\in \I} \frac{4e^6 n'}{(n-n')\binom{n}{n'}}
h_{S, S'_J}
\preceq \sum_{J\in \I} \frac{O(\epsilon)}{\binom{n}{n'}}
h_{S, S'_J},
\]
where we used Lemma~\ref{lem:swap} and the facts that $n' \le \epsilon n$ and $n-n' \ge n-\epsilon n \ge n/2$.
This implies that
\begin{align}
h_S &= \sum_{I\in \I'} \frac{1}{\binom{n}{n'}} h_{S,S_I}
+ \sum_{I\in \I''} \frac{1}{\binom{n}{n'}} h_{S,S_I}
+\sum_{I\in \I'''} \frac{1}{\binom{n}{n'}} h_{S,S_I} \notag\\
&\preceq \frac{1}{\binom{n}{n'}} \sum_{J \in \I} h_{S,S'_J}
+ O(\epsilon \alpha)
+ \frac{O(\epsilon)}{\binom{n}{n'}} \sum_{J \in \I} h_{S,S'_J}, \label{eq:622}
\end{align}
using the derived bound $|\I''|\le \epsilon \alpha |\I|$.
To conclude the proof, it suffices to bound $h_{S,S'_J}$ in terms of $h_{S',S'_J}$.

\begin{lemma} \label{lem:S-Sp}
	Fix a subset $T \subseteq X$ of size $n'$. Then, $h_{S,T} \preceq e^{3 \epsilon} h_{S',T}$.
\end{lemma}
\begin{proof}
	For any $h \in H_T$, let $\lambda_{h,S}$ and $\lambda_{h,S'}$ denote the values of $\lambda_h$ in the definition of $h_{S,T}$ and $h_{S',T}$, respectively. Similarly, define $Z_S$ and $Z_{S'}$. Note the following:
	\[
	\lambda_{h,S}
	= e^{-L_S(h) / \eta}
	\le e^{-L_{S'}(h) / \eta + 1/(\eta n)}
	= \lambda_{h,S'} e^{1/(\eta n)}
	\le e^\epsilon \lambda_{h,S'},
	\]
	where we used the condition $1/(\eta n) \le \epsilon$. The same calculation implies that $\lambda_{h,S'} \le e^\epsilon \lambda_{h,S}$, hence
	$Z_S \ge e^{-\epsilon} Z_{S'}$. Next, since $\A^R_{1/(\eta n), \alpha, H}$ has $1/(\eta n)$-private prediction, it holds that $\varphi_S(h) \preceq e^{1/(\eta n)}\varphi_{S'}(h) \preceq e^\epsilon \varphi_{S'}(h)$. The proof concludes, by plugging $\lambda_{h,S} \le e^\epsilon \lambda_{h,S'}$, $Z_S \ge e^{-\epsilon} Z_{S'}$ and $\varphi_S(h) \preceq e^\epsilon \varphi_{S'}(h)$ in the definition of $h_{S,T}$.
\end{proof} 

The proof of Lemma~\ref{lem:main} now follows from Lemma~\ref{lem:S-Sp} and \eqref{eq:622}:
\[
h_S
\preceq (1 + O(\epsilon)) \frac{1}{\binom{n}{n'}} \sum_{J \in \I} h_{S,S'_J} + O(\epsilon \alpha)
\preceq (1 + O(\epsilon)) e^{3\epsilon} \frac{1}{\binom{n}{n'}} \sum_{J \in \I} h_{S',S'_J} + O(\epsilon \alpha)
= (1 + O(\epsilon)) e^{3\epsilon} h_{S'} + O(\epsilon \alpha).
\]

\section{Proof of Theorem~\ref{thm:sta-lb}} \label{sec:pr-lb}
	By the definition of the VC dimension, there exists a subset $X' \subseteq X$ of size $|X'| = d$ such that the restriction of $H$ to $X'$ contains all binary functions over $X'$, and assume without loss of generality that $X' = \{1,\dots,d\}$.
	 	
	Define a distribution $D_{X'}$ over $X'$ as follows: for any $k \in \{1,\dots,d-1\}$, $D_{X'}(k) = 4\alpha/(d-1)$ and $D_{X'}(d) = 1-4\alpha$. For any hypothesis $h\colon X' \to \{0,1\}$, let $D_h$ be the following distribution over pairs $(x,y)\in X' \times Y$: first, $x \sim D_{X'}$ is drawn, then $y = h(x)$. Let $D_h^n$ denote the distribution over $n$ i.i.d. samples from $D_h$.

	Given $S = ((x_i,y_i))_{i=1}^n$, denote by $S^{\flipk}$ the set obtained from $S$ by flipping all values $y_i$ corresponding to $x_i = k$, namely $S^\flipk = ((x_i, y_i'))_{i=1}^n$ where $y_i' = 1 - y_i$ if $x_i = k$ and otherwise $y_i' = y_i$.
	Let $\#_S(i) := |\{ i \in [n] \colon x_i = k \}|$ denote the number of times that $k$ appears in $S$.
	
	Drawing $h \colon X' \to \{0,1\}$ uniformly at random, we obtain the following:
	
	\begin{align}
		\alpha
		&\ge \E_{h}\E_{S \sim D_h^n}\left[L_{D_h}(\A(S))\right]
		= \E_{h}\E_{S \sim D_h^n}\left[\sum_{k=1}^{d} D_{X'}(k)\left|h(k) - \A(S)(k)\right|\right]\notag\\
		&\ge \frac{4\alpha}{d-1}\sum_{k=1}^{d-1} \E_{h}\E_{S \sim D_h^n}\left[\left|h(k) - \A(S)(k)\right|\right] \notag\\
		&= \frac{4\alpha}{d-1}\sum_{k=1}^{d-1} \E_{h}\E_{S \sim D_h^n}\left[\frac{1}{2}\left|h(k) - \A(S)(k)\right| + \frac{1}{2}\left|(1-h(k)) - \A(S^{\flipk})(k)\right|\right] \label{eq:110}\\
		&\ge \frac{2\alpha}{d-1}\sum_{k=1}^{d-1} \E_{h}\E_{S \sim D_h^n}\left[\left|(2h(k) - 1) + \A(S^{\flipk})(k) - \A(S)(k)\right|\right] \label{eq:111}\\
		&\ge \frac{2\alpha}{d-1}\sum_{k=1}^{d-1} \E_{h}\E_{S \sim D_h^n}\left[|2h(k)-1| - \left|\A(S)(k) - \A(S^{\flipk})(k)\right|\right] \label{eq:112}\\
		&= \frac{2\alpha}{d-1}\sum_{k=1}^{d-1} \E_{h}\E_{S \sim D_h^n}\left[1 - \left|\A(S)(k) - \A(S^{\flipk})(k)\right|\right] \notag\\
		&\ge \frac{2\alpha}{d-1}\sum_{k=1}^{d-1} (1 - \gamma\E_h\E_{S\sim D_h^n}\left[\#_S(k)\right]) \label{eq:118}\\
		&= 2\alpha (1 - \gamma \frac{4\alpha}{d-1}n)\notag,
	\end{align}
	where \eqref{eq:110} follows from a change of measure argument, \eqref{eq:111} and \eqref{eq:112} follow from the triangle inequality and \eqref{eq:118} follows from the fact that $\A$ is $\gamma$-stable. We obtain that $1-4\gamma\alpha n / (d-1) \le 1/2$ which implies that $n \ge (d-1)/(8\gamma\alpha)$ as required.
\section{Conclusions and open problems}
Our work investigates the sample complexity of agnostic PAC learning with two natural and practically-motivated constraints: uniform stability and prediction privacy. We demonstrate the first general techniques that go beyond the naive averaging that has been used in a number of theoretical and practical works \citep{breiman1996bagging,ShwartzSSS10,HammCB16,PapernotAEGT17,papernot2018scalable,DworkFeldman18,BassilyTT:18}. Our results settle the sample complexity of uniformly stable PAC learning (up to logarithmic factors). At the same time, achieving prediction privacy appears to be a significantly more challenging task and there remains a gap between our best upper and lower bounds. Specifically, the best known lower bound for the problem is $\tilde \Omega(d/(\epsilon\alpha) + d/\alpha^2)$ \citep{DworkFeldman18}, whereas our upper bound is $\tilde O(\min\{d/(\epsilon\alpha^2),d^2/(\epsilon\alpha)\} + d/\alpha^2)$. Closing this gap is a natural open problem. 

Another avenue for future research is the computational complexity of learning with private or stable predictions. Our algorithms rely on the exponential mechanism which cannot be implemented efficiently in general and is usually more computationally expensive than empirical risk minimization (used in the averaging-based approaches). Thus designing general techniques whose computational complexity is comparable to that of ERM is a natural open problem.

The focus of this work is the classical distribution-independent PAC model for binary classification. Sample complexity of learning in this framework is well-understood and it has also been used in many prior works on the sample complexity of privacy-preserving learning. This makes it particularly suitable for investigating the sample complexity of private and stable prediction. At the same time, other learning models are likely to be more suitable for understanding of specific practical applications. Thus it is important to investigate the cost of stability and private prediction in other learning models. One example of such cost has been recently demonstrated in the context of learning from long-tailed data distributions \citep{Feldman19:lt}. Namely, it was demonstrated that uniformly stable (or prediction private) algorithms are inherently sub-optimal for such data distributions.

\iffull
\subsection*{Acknowledgements}
\newcommand{\acks}[1]{#1}
\fi
\acks{
We are grateful to Kobbi Nissim and Uri Stemmer for pointing out that the general approach for converting a realizable PAC learning algorithm to an agnostic one from \citep{beimel2015learning} can be used to prove Corollary \ref{cor:sta-to-pri}.
}
\printbibliography

\else

\iffull
\subsection*{Acknowledgements}
\newcommand{\acks}[1]{#1}
\fi
\acks{
We are grateful to Kobbi Nissim and Uri Stemmer for pointing out that the general approach for converting a realizable PAC learning algorithm to an agnostic one from \citep{beimel2015learning} can be used to prove Corollary \ref{cor:sta-to-pri}.
}
\bibliography{vf-allrefs-local,privstab}
\appendix
\section{Notations for the proof}
\paragraph{A different notation for randomized classifiers:}
Recall that $\A_S$ can be viewed as a randomized classifier, namely, $\A_S(x)$ is a random variable. In the formal proof, we use a different way to describe such classifiers. Instead, a randomized classifier is represented by $h \colon X \to [0,1]$, where $h(x)$ represents the probability to classify $x$ as $1$.
This notation is convenient for describing mixtures of hypotheses: if $h$ satisfies that for all $x$, $h(x)$ is a mixture of $h_1(x),\dots, h_N(x)$ with fixed probabilities $p_1,\dots,p_N$, then according to the introduced notation, $h = \sum_{i=1}^N p_i h_i$. Further, the expected population loss equals $L_D(h) = \E_{(x,y)\sim D}[ |h(x)-y|]$.

\paragraph{Comparing hypotheses ($h \preceq \lambda g + \kappa$):} For two randomized hypotheses $h,g\colon X \to [0,1]$ and $\lambda,\kappa \ge 0$, we write  $h \preceq \lambda g + \kappa$ if for any $x \in X$ and $y \in \{0,1\}$, $|h(x)-y| \le \lambda|g(x)-y| + \kappa$. Note that in this notation, the condition of $(\epsilon,\delta)$-private prediction is equivalent to
$$ \A(S) \preceq e^\eps \cdot \A(S') + \delta. $$

\fi
\fi
	
\end{document}